\PassOptionsToPackage{table}{xcolor}
\documentclass[sigconf, nonacm]{acmart}

\AtBeginDocument{%
  }

\setcopyright{none}

\acmConference{Preprint. Under review.}{}{}

\usepackage{booktabs}
\usepackage{tabularx,mathtools}
\mathtoolsset{showonlyrefs}
\usepackage{colortbl}
\usepackage[normalem]{ulem}
\usepackage[inkscapelatex=false]{svg}
\renewcommand{\hat}{\widehat}

\begin{document}

\title{Generative Sequential Notification Optimization via Multi-Objective Decision Transformers}


\author{Borja Ocejo}
\authornote{Equal contribution.}
\email{bocejo@linkedin.com}
\author{Ruofan Wang}
\authornotemark[1]
\author{Ke Liu}
\authornotemark[1]
\authornote{Work done while at LinkedIn, currently at Pinterest.}
\affiliation{
 \institution{LinkedIn}
 \city{Mountain View}
 \country{USA}}

\author{Rohit Patra}
\author{Haotian Shen}
\author{David Liu}
\affiliation{
 \institution{LinkedIn}
 \city{Mountain View}
 \country{USA}}
\author{Yiwen Yuan}
\author{Gokulraj Mohanasundaram}
\author{Fedor Borisyuk}
\author{Prakruthi Prabhakar}
\affiliation{
 \institution{LinkedIn}
 \city{Mountain View}
 \country{USA}}


\renewcommand{\shortauthors}{Ocejo, Wang, Liu, et al.}

\begin{abstract}
Notifications are an important communication channel for delivering timely and relevant information. Optimizing their delivery involves addressing complex sequential decision-making challenges under constraints such as message utility and user fatigue. Offline reinforcement learning (RL) methods, such as Conservative Q-Learning (CQL), have been applied to this problem but face practical challenges at scale, including instability, sensitivity to distribution shifts, limited reproducibility, and difficulties with explainability in high-dimensional recommendation settings.
We present a Decision Transformer (DT) based framework that reframes policy learning as return-conditioned supervised learning, improving robustness, scalability, and modeling flexibility. Our contributions include a real-world comparison with CQL, a multi-reward design suitable for non-episodic tasks, a quantile regression approach to return-to-go conditioning, and a production-ready system with circular buffer–based sequence processing for near-real-time inference.  Extensive offline and online experiments in a deployed notification system show that our approach improves notification utility and overall session activity while minimizing user fatigue. Compared to a multi-objective CQL-based agent, the DT-based approach achieved a +0.72\% increase in sessions for notification decision-making at LinkedIn by making notification recommendation more relevant.
\end{abstract}




\maketitle

\section{Introduction}
Notifications are an important mechanism for delivering relevant and timely information on large-scale platforms such as LinkedIn. Decisions on when and what to send influence both immediate user interactions and long-term platform value. However, optimizing notification policies presents unique challenges: high notification frequency can lead to fatigue, while overly restrictive approaches may limit delivery of useful information. Achieving an appropriate balance requires models that account for sequential user interactions and long-horizon objectives.

Historically, notification decision-making has relied on supervised learning models trained to predict short-term objectives, such as click-through rate (CTR) or open rate ~\cite{wang2024limamlpersonalizationdeeprecommender}, combined with constrained multi-objective optimization ~\cite{gao2018near}. While effective in some settings, focusing narrowly on short-term metrics can limit long-term outcomes ~\cite{wang2024retentivedecisiontransformeradaptive}.

Recent work has shown that reinforcement learning (RL) can directly optimize notification policies for cumulative measures of platform value, enabling more holistic decision-making. At LinkedIn, offline RL approaches such as Conservative Q-Learning (CQL) ~\cite{kumar2020conservativeqlearningofflinereinforcement} have been deployed to support notification send/drop decisions, improving notification relevance while respecting fatigue signals ~\cite{prabhakar2022multiobjectiveoptimizationnotificationsusing}. Despite their benefits, value-based offline RL methods based on temporal difference (TD) learning ~\cite{sutton2018reinforcement} can be challenging to operate at scale, often exhibiting training instability, reproducibility issues, and limited explainability.

The Decision Transformer (DT) paradigm ~\cite{chen2021decisiontransformerreinforcementlearning} has emerged as a stable and flexible alternative, leveraging return-conditioned sequence modeling to combine the strengths of supervised learning and reinforcement learning. DT reframes policy learning as supervised modeling conditioned on desired long-term outcomes, improving reproducibility and interpretability compared to value-based methods.

In this work, we present the design and deployment of a DT-based solution for notification decision-making at LinkedIn. Key innovations include: (1) a quantile regression method for learning state-dependent return-to-go prompts, enabling multi-objective optimization that incorporates fatigue and outcome constraints; and (2) a circular database buffer–based infrastructure for production-scale transformer inference. Our system delivered a 0.72\% increase in user sessions while maintaining notification relevance and managing fatigue, demonstrating how sequence models can support long-term value in real-world settings.

\section{Related Works}
Traditional notification systems often rely on supervised models ~\cite{gao2018near,yuan2022statetransitionmodelmobile,zhao2018notification} that predict immediate user responses and apply empirically determined thresholds to control message frequency and spacing. While effective for short-term response metrics, such systems can focus too narrowly on immediate outcomes, sometimes leading to high notification volume and user dissatisfaction.  

To address this, model-free offline RL methods based on fitted value or action-value functions have been widely studied in recommendation systems ~\cite{chen2023deep,ie2019slateq,prabhakar2022multiobjectiveoptimizationnotificationsusing,chen2022off,levine2020offline}. Modeling  the long-term objectives within RL framework helped achieve coordinated notification timing and improved outcomes with fewer notifications~\citep{obrien2022isendnotificationoptimizing,prabhakar2022multiobjectiveoptimizationnotificationsusing}. Similarly, ~\cite{ji2024timtemporalinteractionmodel} used historical interaction data to predict dynamic click-through rates throughout the day, resulting in improved response rates with reduced disturbance. These approaches exploit the Markov property to decompose long-horizon decision-making into single-step learning problems. Despite their success, scaling these methods for production use is challenging due to training instability and sensitivity to hyperparameters, which can affect reproducibility ~\cite{kumar2020conservative,prabhakar2022multiobjectiveoptimizationnotificationsusing,obrien2022isendnotificationoptimizing}. An alternative perspective is to treat RL as a sequence generation problem, aiming to produce action sequences leading to high returns. Transformers~\cite{vaswani2023attentionneed} naturally align with this paradigm because of their capacity to model long-term dependencies and scale efficiently with data and compute ~\cite{janner2021offline,chen2021decisiontransformerreinforcementlearning}. While supervised RL approaches such as DT lack formal optimality guarantees, this limitation is less critical in non-robotic tasks like recommendation systems, where iterative policy improvements are prioritized over finding globally optimal policies.

In this framework, RL is reformulated as conditional imitation learning or behavior cloning ~\cite{emmons2021rvs}, with conditioning variables often based on returns ~\cite{kumar2019reward,chen2021decisiontransformerreinforcementlearning,chen2023deep,lee2022multi,liu2023constrained} or goals ~\cite{ghosh2019learning,codevilla2018end}. Our work belongs to the return-conditioned supervised learning (RCSL) family. A known limitation of RCSL is that arbitrary conditioning values can lead to poor performance ~\cite{brandfonbrener2022does}, motivating techniques that learn or plan these conditioning values directly ~\cite{lee2022multi,emmons2021rvs}. Our approach is also connected to hierarchical RL and hierarchical Decision Transformers (HDT), which address compounding errors in long-horizon sequential decision-making. HDT approaches ~\cite{correia2023hierarchical,badrinath2023waypoint,ma2023rethinking} often use separate models for goal learning, whereas our method and ~\cite{clinton2024planning,lee2022multi,gao2025generative} adopt a unified architecture for planning and control. Unlike ~\cite{clinton2024planning}, we model conditional return-to-go distributions as planning targets, providing increased flexibility and interpretability compared to point estimates.

\section{Problem Setup}
We operate a nearline system for notification decision-making that processes hundreds of notification candidates per user for millions of users. Notifications are generated continuously at arbitrary times throughout the day. To process them efficiently, we aggregate all incoming notifications into batches and evaluate them in scheduled cycles at a fixed cadence. At any given evaluation cycle, only a single notification can be sent to a user, either as an app badge update or a push notification. A multitask ranking model scores all available notifications for a user and selects the top-ranked item. This top item is then evaluated by a decision-making agent, which, at each time step ${t}$, observes the current state $\mathbf{s_t}$, selects an action ${a_t}$, and receives a vector of rewards $\mathbf{r_t} \in \mathbb{R}^{|\mathcal{R}|}$. The objective is to learn a single optimal policy, parameterized by $\mathbf{\theta}$, that maximizes the agent’s total expected discounted return over time $\mathbf{R_t}$. We provide a detailed explanation of how we define states, actions, and rewards.

\textbf{States.} $s$ denotes a state in the state space $\mathcal{S}$. We use a carefully engineered set of features to represent the state, including the number and types of notifications sent in the past x hours, user profile attributes, time since last visit, historical visit rates, notification content understanding and other contextual signals about the user and the candidate notification.

\textbf{Actions.} $a$ represents an action in the discrete action space $\mathcal{A}$, which includes: send as a badge, send as a push, and don’t send. For any given notification candidate, the set of eligible actions may include one, a few, or all of the above.

\textbf{Rewards.} $\mathbf{r}_t \in \mathbb{R}^{|\mathcal{R}|}$ represents a vector of multi-objective rewards in the reward space $\mathcal{R}$. DT must balance multiple competing objectives to derive an optimal policy. The first class of rewards comes from predicted measures of notification value, such as the likelihood of clicking an in-app notification or a push notification, depending on the action taken. While prediction models introduce inherent bias, they also help reduce the variance in offline learning, enabling a more favorable bias–variance tradeoff and thus more efficient policy optimization. These predicted reward signals are derived from existing, well-optimized models already deployed for ranking and top-notification selection problems. The second class of rewards captures actual user visits that occur between consecutive states. To manage user fatigue and limited attention, adaptive volume penalty rewards are also incorporated, ensuring optimal pacing of notifications.

This framework allows us to treat notification delivery as a long-horizon, multi-objective control problem, requiring the agent to reason about sequential dynamics, balance competing objectives, and generalize effectively across hundreds of millions of users in a highly dynamic environment.

\section{Methodology}

\begin{figure*}[ht]
    \centering
    \includegraphics[width=0.85\textwidth]{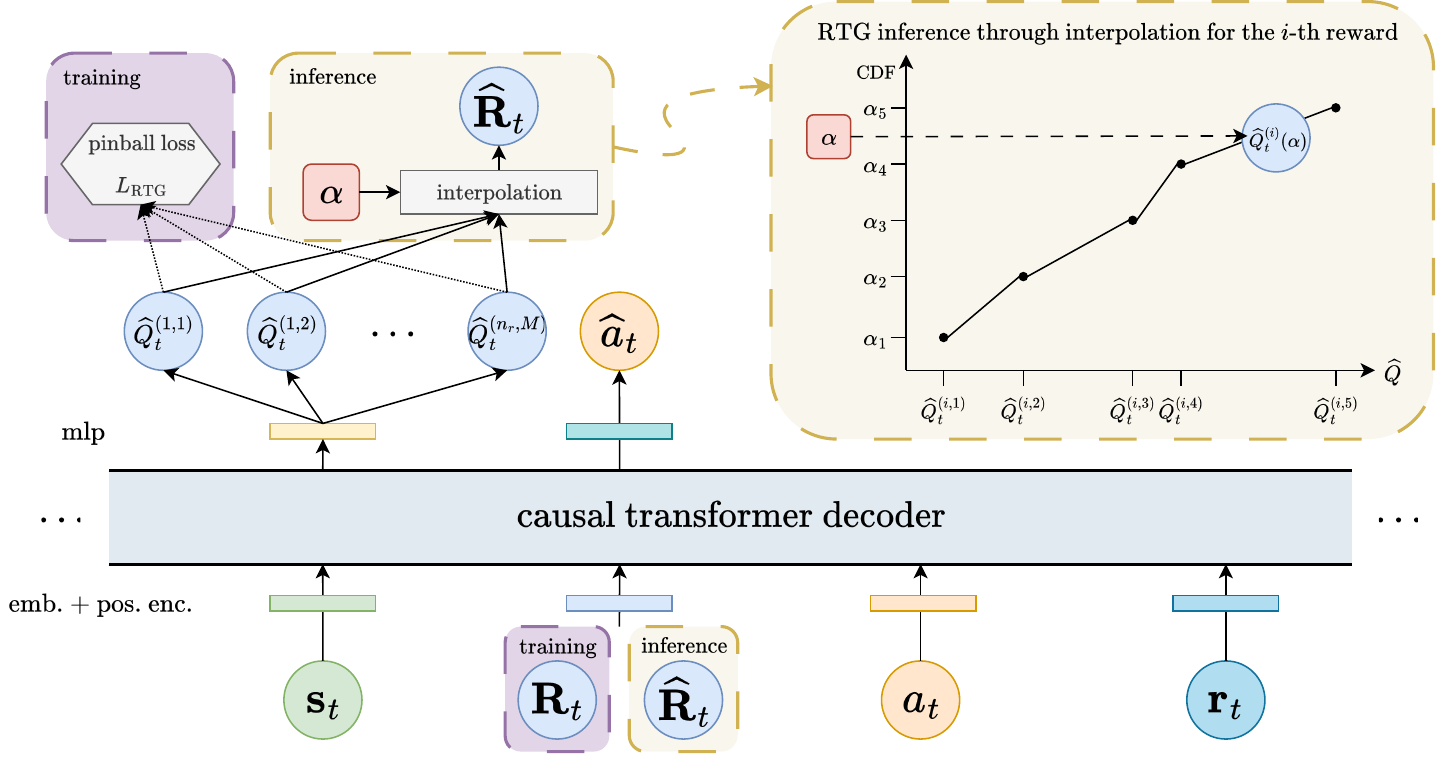}
    \caption{Model Architecture: During training, DT learns RTGs at a fixed set of quantile levels \smash{$\{\alpha_j\}_{j=1}^M$} using quantile regression and learns action given the ground truth RTG $\mathbf{R}_t$. During inference, for a target quantile level \smash{$\alpha$}, DT first predicts \smash{$\mathbf{\hat{R}}_t$} using linear approximation from the learned quantiles \smash{$\hat{Q}_t$}, then autoregressively predicts action \smash{$\hat{a}_t$} given the predicted \smash{$\mathbf{\hat{R}}_t$.}}
    \label{fig:banner}
\end{figure*}

\subsection{Decision Transformer}
At the heart of our framework is a DT model, which casts offline RL as a sequence modeling problem. We adopt the trajectory representation in ~\cite{lee2022multi},
\begin{equation}
    \tau = \left( \mathbf{s}_1, \mathbf{R}_1, a_1, \mathbf{r}_1, \dots, \mathbf{s}_T, \mathbf{R}_T, a_T, \mathbf{r}_T \right)
\end{equation}
where $\mathbf{R}_t$ is the return-to-go (RTG).

Unlike typical RL applications such as game playing or robotic control where episodes have natural termination conditions, notification optimization operates in a continuous, infinite-horizon setting.
We reformulate our task into finite-horizon episodes by sampling trajectories of length $T+H$ from the continuous user interaction stream, where $T$ represents DT's context length and $H$ denotes the look-ahead horizon for reward computation.
For each timestep $t$ within a trajectory, we calculate the RTG as $\mathbf{R}_t = \sum_{l=0}^{H}\gamma^{l}\mathbf{r}_{t+l}$.
The look-ahead horizon $H$ provides a trade-off between capturing long-term behavioral patterns and keeping computational complexity manageable, while the discount factor $\gamma$ places higher weight on near-term user responses relative to distant future outcomes.

We opt to preserve the multi-dimensional reward structure as a vector $\mathbf{r}_t \in \mathbb{R}^{|\mathcal{R}|}$ rather than combining individual reward components through predefined preference weights.
This design choice allows DT to learn an implicit functional form for reward combination, potentially discovering more sophisticated relationships between reward components than human-engineered linear combinations could capture. Additionally, maintaining separate reward dimensions allows granular control during inference, where we can independently adjust different reward components to tune the model's behavior. This flexibility is critical for dynamically adapting notification decision-making, enabling adjustments to evolving business priorities and optimizing for different reward objectives across user cohorts over time.

At each timestep, we obtain a set of eligible actions (EAS) as a tensor. We integrate the EAS embedding with the hidden representation used for action prediction via a linear transformation, followed by a multiplicative interaction between the two transformed vectors. This enables the network to incorporate eligibility constraints directly into the action prediction process, effectively conditioning its outputs on the subset of admissible actions. The action prediction is learned via trajectory aggregate cross-entropy loss 
\begin{align}
L_{\text{action}}:= \sum_{t=1}^T \text{CrossEntropyLoss}(a_t, \hat{a}_t).
\end{align}
We did not train DT to explicitly predict future states or rewards; in other words, the training objective focused solely on actions and RTG predictions.

\subsection{Return Prompt Modeling}
One of the key aspects of the decision transformer is the conditioning on target future returns, or return-to-go. The original implementation defines return-to-go at time $t \leq T$ as
\begin{equation}
    R^{'}_t = R^{'}_1 - \sum_{i=1}^{t}r_i
    \label{eq:manual-rtg-update}
\end{equation}

where $r_i$ is the observed reward at time $i$ and ${R}^{'}_{1}$ is the (non-discounted) RTG for the \emph{entire} trajectory. In this implementation, a desired RTG prompt is set once by selecting the highest achieved $R^{'}_1$ and sequentially updates the prompt using~\eqref{eq:manual-rtg-update} for time $t \leq T$. This manual update poses a few challenges for our notifications decision making system:
\begin{itemize}
    \item Some of the rewards (e.g., CTR) are not realized before the next actions need to be predicted. This makes the manual update in~\eqref{eq:manual-rtg-update} infeasible.
    \item If the prompt is outside a reasonable range, the model's expected return based on the predicted action might be drastically different from the desired return. We need a mechanism that derives the prompt at inference time to ensure that return-to-go at time $t$ does not deviate too much from the underlying distribution of $R_t$.
\end{itemize}
We propose a way to model return-to-go within our decision transformer in section ~\ref{quantile-reg} to address these challenges.

\subsubsection{Training Procedure} \label{quantile-reg}
When using the manual update, we need a way to determine a reasonable $R_1$ as a starting point in inference - something that can be particularly tricky in real world settings. Rather than depending on manual methods of determining this value, we propose that the model learn the distribution of $R_t$. In particular, we want to learn likely expert-level returns using the existing history ~\cite{lee2022multi}.

Rather than modeling the return-to-go updates as in the heuristic updates above, we want to predict $R_t$ at inference time, using the history up to time t. Let $\tau_{<t}$ be the history before timestep $t$,
\begin{equation}
\tau_{<t} = (s_1, \hat{R}_1, a_1, r_1, ..., s_{t-1}, \hat{R}_{t-1}, a_{t-1}, r_{t-1}).
\end{equation}
At each time $t$, the model input is $[\tau_{<t}, s_t]$, the history before $t$ and the current state $s_t$. We aim to learn the conditional quantile of the $i$-th return-to-go; ${Q}^*_{R^{(i)}}(\alpha; \tau_{<t}, s_t )$ for a set of given quantile levels $\{\alpha_j\}_{j=1}^M \subset (0, 1)$.

Suppose we have a predetermined set of $M$ quantiles $\{\alpha_j\}_{j=1}^M \subset (0, 1)$. Let $\mathbf{h}_t^R \in \mathbb{R}^{d_{model}}$ be the hidden representation of the return-to-go token at time $t$. The hidden representation is passed through a following two layer MLP
\begin{align}
\hat{Q}_t^{flat} = \text{MLP}_R(\mathbf{h}_t^R) \in \mathbb{R}^{n_rM},
\end{align}
where $n_r := |\mathcal{R}|$. The output vector $\hat{{Q}}_t^{flat}$ is reshaped to represent the matrix of predicted quantiles $\hat{Q}_t \in\mathbb{R}^{n_r \times M},$ where

\begin{equation}
    \hat{Q}_{t}^{(i, j)} :=\hat{Q}_{R^{(i)}}(\alpha_j; \tau_{<t}, s_t )  \approx Q^{*}_{R^{(i)}} (\alpha_{j};  \tau_{< t}, s_t),
\end{equation}
where $i \in \{ 1, ..., n_r \}$ represents the reward index and $j \in \{ 1, ..., M \}$ represents a quantile index. Parameter updates during training for the return-to-go (RTG) prediction model are driven by the following trajectory aggregate loss:
\begin{align}\label{eq:sum_loss}
    L_{\text{RTG}} = \sum_{t=1}^T\sum_{i=1}^{n_r} \sum_{j=1}^M \rho_\alpha\big(\hat{Q}_{t}^{(i, j)}, R_t^{(i)}\big),
\end{align}
where $\rho_\alpha$ is pinball loss
\begin{equation}
    \rho_{\alpha} (\hat{y}, y) = \begin{cases}
        \alpha \cdot (y - \hat{y}), & \text{if } y \geq \hat{y} \\
        (\alpha - 1) \cdot (y - \hat{y}), & \text{if } y < \hat{y}.
    \end{cases}
\end{equation}
We use the RTG and action losses in a linear combination to minimize the total DT loss $L_{\text{DT}}$
\begin{equation}
    L_{\text{DT}} = L_{\text{action}} + \lambda \cdot L_{RTG},
\end{equation}
where $\lambda$ is a hyperparameter that represents the weight of the return-to-go loss relative to the action loss.

\subsubsection{Inference Procedure}

Although the model learns a limited set of $M$ quantile levels, we wanted the flexibility to predict quantile levels that are not part of the set $\{\alpha_j\}_{j=1}^M$ without retraining the model. If $\alpha$ is the target return-to-go quantile level for the $i$-th reward, we use the linear approximation below to estimate the RTG inference prompt
\begin{align}
\hat{\mathbf{R}}_t^{(i)} := \hat{Q}_t^{(i)}(\alpha) &:= \lambda_\alpha \, \hat{Q}_{t}^{(i,l)} +  (1- \lambda_\alpha) \, \hat{Q}_{t}^{(i,u)} \label{eq:quantile-approx} 
\,\text{with}\,
\lambda_\alpha := \frac{\alpha_u - \alpha}{\alpha_u - \alpha_l}, 
\end{align}
where $u := \min\left\{ j : \alpha_j > \alpha \right\}$, and $l := \max\left\{ j : \alpha_j < \alpha \right\}$.

This return modeling approach means that we can directly "ask", or prompt, for a desired $\alpha$ quantile level return at inference time without having to test several arbitrary and difficult-to-interpret prompt values for $R_t$. In addition, the linear approximation lets us prompt quantile levels that were not part of the initial training set. The added flexibility enables us to select $\alpha$ levels that enhance the user experience.


\section{Production Setup}

\subsection{Background}
Within the LinkedIn notifications ecosystem, the core serving infrastructure is powered by the Air Traffic Controller (ATC) ~\cite{linkedin2021atc}, a nearline stream processing job built on Apache Samza. ATC optimizes the notification experience for each user by making near-real-time delivery decisions. It integrates tightly with internal LinkedIn platforms such as Venice (a distributed key-value store) ~\cite{linkedin2022venice} and our in-house feature computation and retrieval engine to enable Decision Transformer (DT) inference at scale. The system handles an average throughput of 100K queries per second (QPS), with peaks reaching up to 150K QPS during busy hours.

\subsection{Sequence Persistence via Circular Buffer with Partial Update}
\begin{figure}[ht]
    \centering
    \includegraphics[width=\columnwidth]{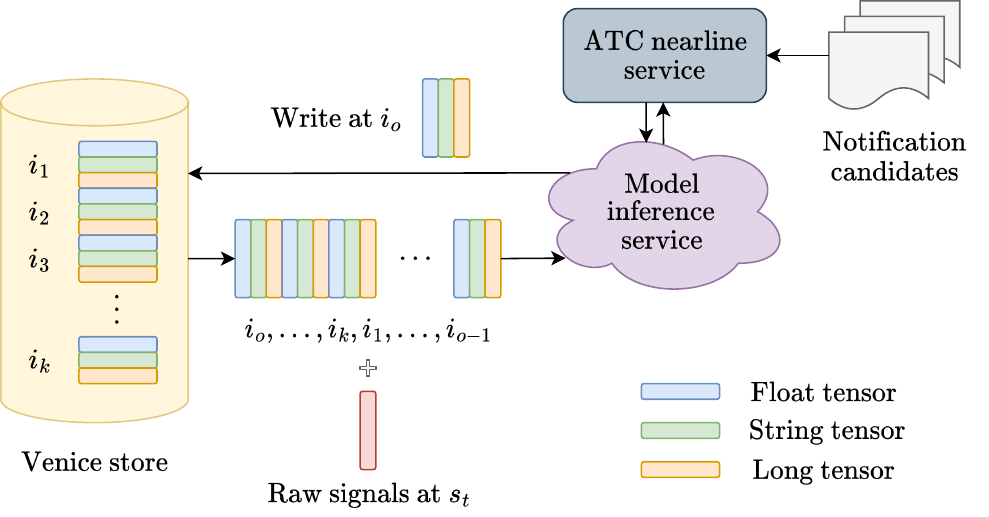}
    \caption{Production architecture: Notification candidates are processed by the ATC nearline service, invoking a model cloud service, that temporally orders user-specific float, long, and string tensors via a circular buffer, performs inference over the sequence, updates tensors for the new timestep, and returns scored candidates.}
    \label{fig:dt_inference}
\end{figure}
At each timestep $t$, ATC receives a set of notification candidates for inference and calls the model cloud inference service ~\cite{10.1145/3627673.3680091} with these candidates. The inference service retrieves both (i) historical sequence features and (ii) current state feature tensors for the user from remote Venice stores. For each user, the sequence Venice store stores $K$ tuples of Float, Long, and String tensors as an Avro schema, enabling Transformer-based inference with up to sequence length 16. Conceptually, these tensors are organized as a fixed-length circular buffer, where the oldest timestamp resides at index $o$, and indices $o, ..., K, 1, ..., o-1$ represent the true chronological order of events. Although the circular buffer’s physical storage order is not strictly chronological, each entry contains a persisted timestamp, enabling correct ordering reconstruction at inference time. This design minimizes storage cost while supporting sequence-aware modeling and efficient updates.

The model leverages this historical context and the current state to score utility models, select the top-1 notification, and compute decision transformer predictions. After inference, the model constructs a new tensor tuple (FLS tensor: Float, Long, and String tensors) representing the current timestep and writes it back to Venice using its partial update API, ensuring only incremental changes are persisted. An internal index pointer is used to overwrite the oldest entry in a round-robin manner. This write-back step allows rewards, actions, and features to evolve sequentially with every inference call, minimizing overhead while preserving temporal structure. Non-model-based external rewards are ingested as features and appended to the FLS tensor before being written back to Venice.

Additionally, the model supports selective use of history: using a string key in the String tensor, parts or all of the historical sequence can be ignored. This capability is used when certain utility models in the current state have changed from those used in the historical sequence, or to bootstrap a new model variant without relying on older model-specific history.

To ensure data hygiene and operational agility, we maintain two offline Apache Airflow workflows:
\begin{itemize}
    \item \textbf{Empty Push Flow}: Clears all records in the store, typically used during major model version upgrades.
    \item \textbf{TTL Repush Flow}: Periodically deletes records older than 14 days to maintain data freshness and quality.
\end{itemize}
To stay within our storage quota, we enabled \textbf{zstd compression} ~\cite{zstd} on the store. zstd is a fast lossless compression algorithm that achieved a 5x compression ratio in our production setup.

\subsection{Cloud-based Model Serving}
We adopt a decoupled architecture that separates model inference from application logic. This enables us to scale model serving independently of Samza and ATC, which is especially beneficial as model complexity grows. Horizontal and vertical scaling of model inference is achieved in a cost-effective manner.

Each model runs in its own isolated tenant environment, enabling tailored configuration of hardware and performance parameters. For example, simpler models may run on standard CPUs, while more complex variants leverage GPUs or high-performance CPU SKUs. This per-model isolation also facilitates safe experimentation with multiple models in parallel, each with its own resource footprint and roll-out schedule.

\section{Experimental Results}
In this section, we present the performance of DT against its CQL baseline from both offline and online experiment results. We also describe several crucial metrics in the notifications ecosystem, which help evaluate the contributions of different modeling technologies and architectural changes.

\subsection{Experiment Setup}
We collect $\epsilon$-greedy randomized data by deploying the baseline CQL policy to 2\% of randomly selected LinkedIn users over one week. For each user, we record the full sequence of notification interactions, resulting in trajectories of the form $(s_1, R_1, a_1, r_1, \ldots, s_T, R_T, a_T, r_T)$. Users are randomly partitioned into 70\% for training and 30\% for validation. Because user activity patterns vary, episode lengths differ across users.

To prepare model inputs for the Decision Transformer, we segment each user’s interaction history into \textit{trajectories} of fixed length $T+H$, where $T$ is the historical context length and $H$ is the look-ahead horizon. Each trajectory contains $T$ consecutive past steps followed by $H$ future steps, providing both sufficient context and a horizon window for return optimization. For users with long interaction histories, overlapping trajectories are generated by sliding the context window through time.

During training, we monitor two task-specific metrics: \textit{action accuracy} for next-action prediction and \textit{pinball loss} for return-to-go estimation. For selected policies, we perform online rollouts to 2\% of LinkedIn users and evaluate key operational metrics:

\begin{itemize}
\item \textbf{Sessions}: A session is defined as a series of full-page views by a user on the same device, separated by at least 30 minutes. This metric reflects user activity on the platform~\cite{prabhakar2022multiobjectiveoptimizationnotificationsusing}.
\item \textbf{Notification Volume}: The total number of notifications delivered, including push and in-app notifications.
\item \textbf{Notification CTR}: The click-through rate of notifications sent within a day, a key indicator of notification relevance and quality.
\end{itemize}

A superior policy attains a more favorable balance between Sessions and Volume without compromising CTR. Ideally, we observe either an increase in sessions at a comparable notification volume, reflecting improved efficiency, or neutral session counts with a reduction in overall notification volume.

\subsection{Offline Results}

We trained the DT models on the training split and evaluated their performance on the validation set. The action prediction task is formulated as a three-class classification problem and for the return-to-go prediction task, we employed quantile regression to estimate the $0.25$, $0.5$, and $0.75$ quantiles. The best DT policy achieves an accuracy of $96.7\%$ with a pinball loss of $0.358$ across these targets. 


\subsection{Online A/B Test Results}

We conducted a series of online A/B experiments to evaluate DT based notification policies deployed at LinkedIn. In these experiments, 2\% of randomly selected users were served notifications generated by candidate DT policies, with the existing CQL–based policy used as the baseline. Each experiment ran for one week. Table~\ref{tab:ab_results} summarizes relative percentage changes for three key metrics: User Sessions, Notification Volume, and CTR. The first row reports the overall results from the best-performing DT policy compared to the CQL baseline. The following rows present incremental changes from individual model enhancements, including: (1) a basic DT configuration with historical context length set to 1 and a constant return prompt at the cohort level, (2) the addition of learned return-to-go prompts derived from state features, and (3) an increase in historical context length.

We use a richer state representation compared to the CQL baseline. Incorporating these signals into a CQL agent with an enlarged network did not yield stable or effective policies, likely due to the training instabilities commonly observed in TD learning. For the basic DT setup, we use a constant prompt corresponding to the 70th quantile of the return-to-go distribution for each reward type, computed at the cohort level. For experiments with learned prompts and extended historical context, we provide a set of quantiles $\{\alpha_j\}_{j=1}^M$ centered around 0.7 and use prompt tuning to determine the optimal prompt for each configuration.

\begin{table}[h]
\centering
\caption{Online A/B test results for DT notification policies. The first row shows overall DT vs. CQL gains; others report incremental impact. NSS: not statistically significant.}
\begin{tabularx}{\linewidth}{l *{3}{>{\centering\arraybackslash}X}}
\toprule
\textbf{Policy/Enhancement} & \textbf{Sessions (\%)} & \textbf{Notif. Vol. (\%)} & \textbf{CTR (\%)} \\
\midrule
\rowcolor{gray!20}
\textbf{Overall DT vs. CQL} & \textbf{0.72} & \textbf{-1.68} & \textbf{NSS} \\
Basic DT (Context Len = 1)              & 0.51  & -0.31 & -0.66 \\
Learned Prompts        & NSS     & -1.18 & 0.43 \\
Longer Context Length        & 0.21  & -0.19 & NSS  \\
\bottomrule
\end{tabularx}
\label{tab:ab_results}
\end{table}

Overall, the DT policy achieved a 0.72\% increase in sessions while reducing notification volume, with no statistically significant change in the guardrail metric (CTR) relative to the CQL baseline. These findings indicate that the proposed approach can effectively optimize multi-objective notification policies at production scale. The observed improvements result from a combination of modeling and system-level enhancements: incorporating learned return-to-go prompts allowed the policy to maintain session activity with fewer notifications, while extending the historical context length provided additional gains in session metrics.

\subsection{Ablation Study}
In this subsection, we share observations and insights from experimenting with different training and serving configurations.

\subsubsection{Impact of Context Length}
We evaluate the effect of input context length on both model performance and serving efficiency. Specifically, we experiment with context lengths of 1, 2, 4, and 8 steps. We observe that increasing the context length improves action prediction accuracy due to richer temporal context. However, longer sequences also incur higher serving costs, as more historical data must be retrieved and processed per inference request.

We compute the average action prediction accuracy on the evaluation dataset over five different training random seeds as well as the standard deviation. We quantify serving cost by measuring the number of CPU instances required to handle 1\% of LinkedIn public traffic. Table~\ref{tab:context_ablation} shows the relative changes in both action prediction accuracy and serving cost, normalized to the context length of 1.

\begin{table}[h]
\centering
\caption{Relative impact of context length on model accuracy and serving cost with standard deviations.}
\label{tab:context_ablation}
\begin{tabularx}{\linewidth}{c X X}
\toprule
\textbf{Context Length} & \textbf{Action Prediction Accuracy} & \textbf{Serving Cost} \\
\midrule
1 & - & - \\
2 & +0.094\% {($\pm$ 0.023\%)} & +10\% \\
4 & +0.194\% {($\pm$ 0.011\%)} & +25\% \\
8 & +0.202\% {($\pm$ 0.029\%)} & +60\% \\
\bottomrule
\end{tabularx}
\end{table}

To balance model quality and system efficiency, we selected a context length of 4 for production deployment. This configuration provides a strong accuracy boost over shorter contexts, while maintaining serving cost within acceptable limits.

\subsubsection{Enhancing Action Prediction with Richer Context}

In the traditional auto-regressive paradigm, the action token is predicted using only the hidden state corresponding to the return-to-go (R) token. In our experiments, we also tried concatenating the hidden states of both the state ($s$) and return ($R$) tokens to predict the action ($a$). This richer input led to a notable improvement in action prediction accuracy during offline evaluation.

\begin{table}[h]
\centering
\caption{Enhanced context input boosts action prediction accuracy}
\begin{tabular}{lc}
\toprule
\textbf{Input Representation} & \textbf{ Accuracy} \\
\midrule
Return token only ($R$) & - \\
State + Return ($s + R$) & +0.064\% {($\pm$ 0.031\%)} \\
\bottomrule
\end{tabular}
\end{table}

\subsection{Prompt Tuning Case Study}
The quantile regression setup offers flexibility in specifying target outcomes for each reward objective. We refer to this process as \textbf{prompt tuning}, wherein the target quantiles $\{\alpha_j\}_{j=1}^M $ are tuned to yield distinct notification policies with varying behaviors. In this section, we present a case study illustrating how prompt tuning on a specific reward (CTR) can be used to directly influence the resulting policy.

\begin{figure}[htbp!]
    \centering
    \vspace{-.1in}
    \includegraphics[width=0.4\textwidth]{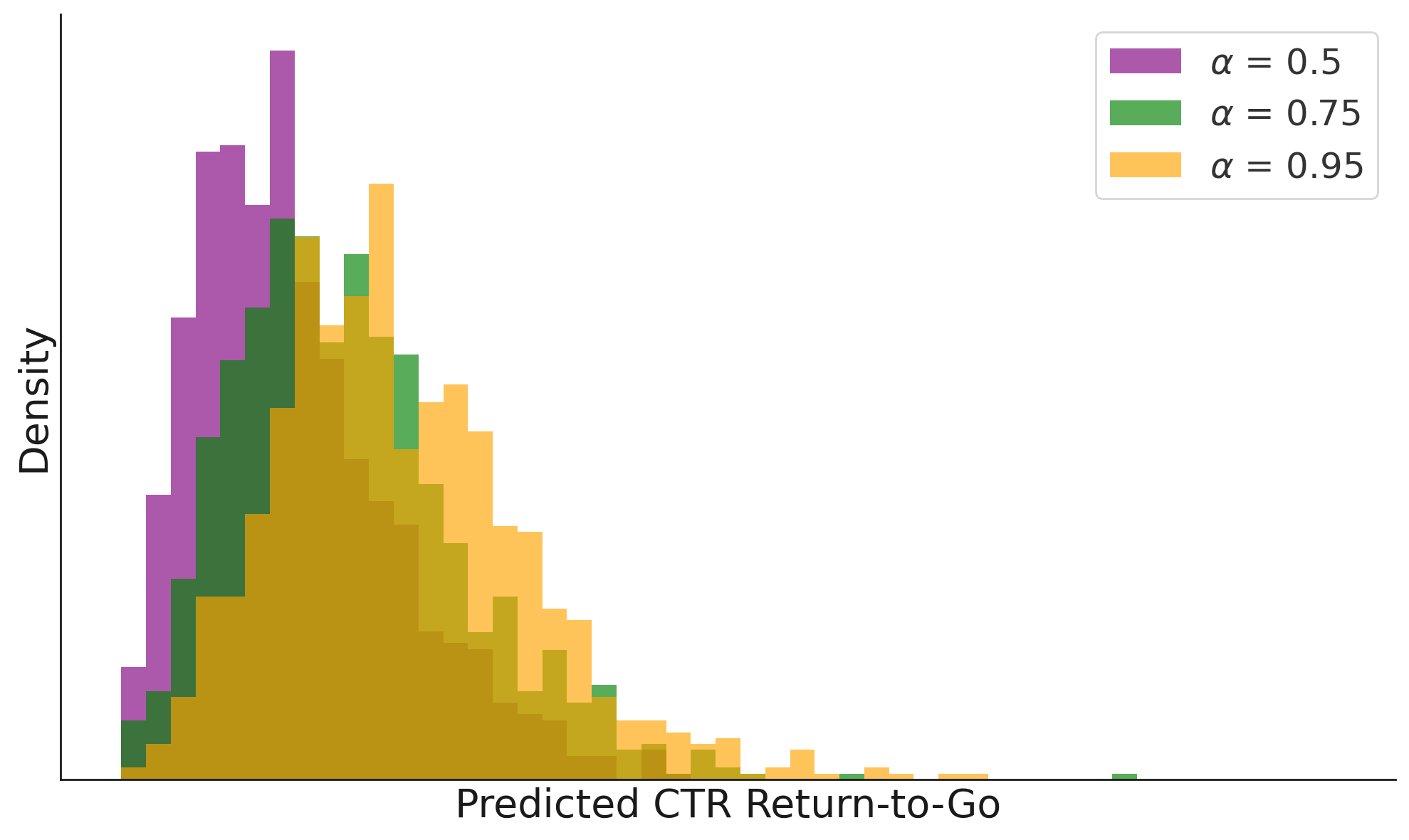}
    \caption{Predicted return-to-go for CTR at different target quantiles}
    \label{fig:ctrrtgoffline}
    \vspace{-1.0em}
\end{figure}

Figure~\ref{fig:ctrrtgoffline} illustrates the distribution of learned return-to-go values for the CTR reward under different quantile prompts ($\alpha = 0.5, 0.75, 0.95$). As the target quantile increases, the distribution shifts systematically to the right. This demonstrates the effect of prompt tuning: by adjusting the quantile parameter, the reward-seeking behavior of the policy can be modulated, enabling the generation of notification strategies that reflect different reward priorities.




\begin{figure}[htbp!]
    \centering
    \includegraphics[width=0.4\textwidth]{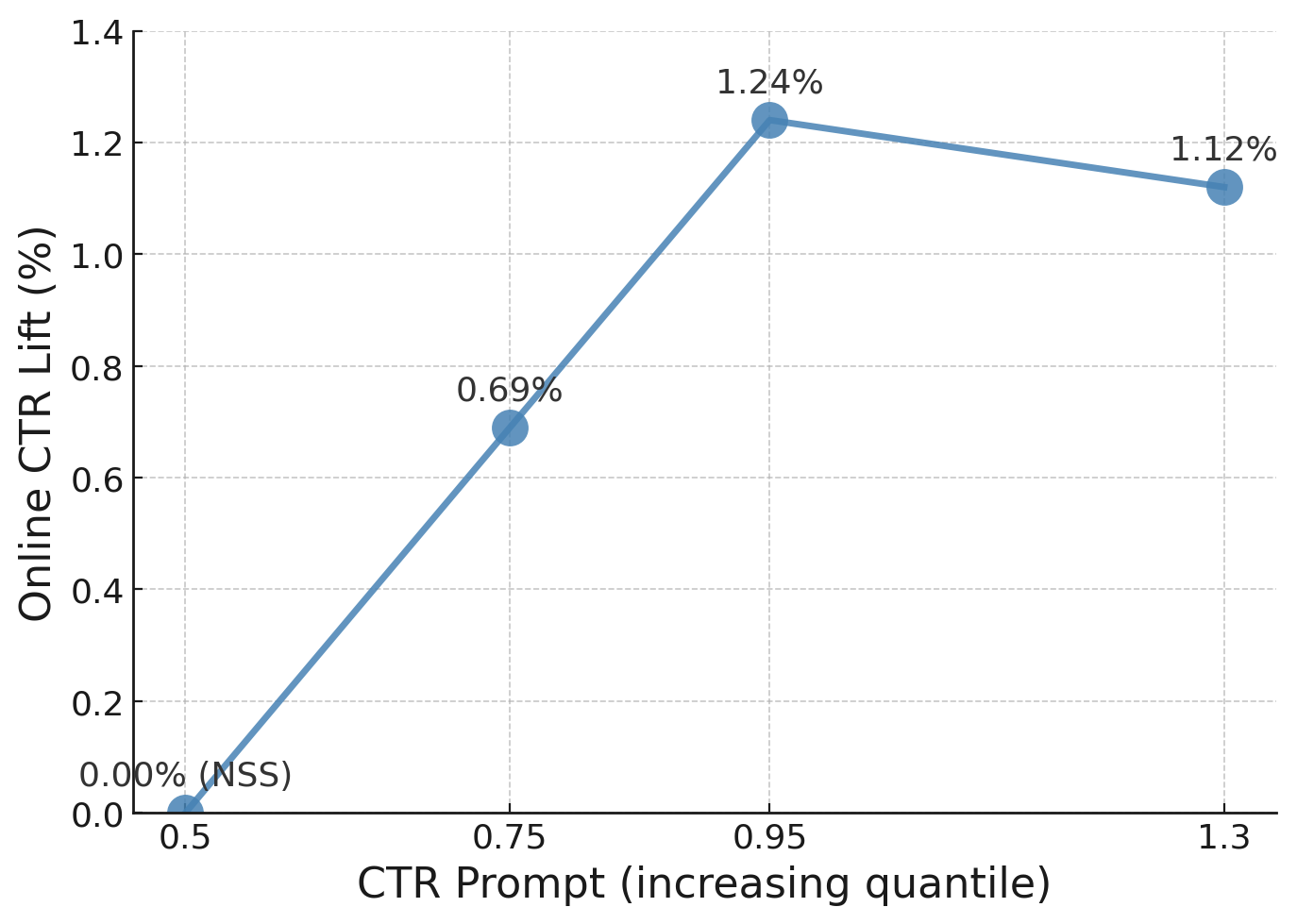}
    \caption{Effect of CTR Prompt Tuning on Online CTR}
    \label{fig:onlinectr1}
    \vspace{-1.0em}
\end{figure}

CTR serves as an indicator of the relevance and quality of delivered notifications. Increasing the return-to-go value associated with the CTR reward is expected to raise online CTR. As shown in Figure~\ref{fig:onlinectr1}, online CTR consistently improves as the quantile of the CTR prompt increases; each point represents the outcome of an independent online A/B test under a different prompt configuration. These results demonstrate that prompt tuning provides fine-grained control over policy objectives related to notification quality and delivery strategies, enabling systematic exploration of the decision space and identification of effective policy settings.


\section{Deployment Lessons}
Deploying DT–based notification policies at LinkedIn has led to several practical insights that may inform future work.

\textit{Feature Scalability and Training Stability:} A significant limitation of value-based offline RL methods such as CQL is their sensitivity to high-dimensional feature sets. In our previous experiments, attempts to expand the state representation feature space often led to pronounced training instability and convergence challenges. In contrast, DT maintained stable training and achieved significant performance gains as the feature set expanded, enabling a more than tenfold increase in feature dimensionality and richer user and context representations that were infeasible with CQL. In real-world recommender systems, the ability to rapidly iterate on the state space and enhance feature representations is critical for sustaining long-term performance gains.

\textit{Robustness and Model Efficiency:} DT also proved substantially more robust than CQL. With CQL, training different policies on the same data could result in unpredictable and widely varying behaviors, complicating both evaluation and online iteration. The DT framework produced much more consistent results across training runs and policy configurations, which enabled us to reduce the number of required candidate models by over $3\times$ and accelerate our experimentation cycle, boosting developer productivity.

\textit{Improved Policy Tuning Flexibility and Interpretability:} The return-conditioned nature of DT made policy behaviors more interpretable and tunable. Prompt tuning with quantile regression enabled direct and transparent adjustment of policy objectives, allowing us to align system-level outcomes with business goals more flexibly in a personalized manner than was possible with value-based RL. This has been repeatedly observed and confirmed from our day-to-day DT iterations.

Overall, the move to Decision Transformers not only led to superior online performance, but also fundamentally improved the engineering and operational aspects of developing and maintaining sequential decision-making models at scale.

\section{Conclusions}
In this work, we present a Decision Transformer–based approach for generative, sequential decision making applied to multi-objective notification optimization. We frame notification decision making as a reinforcement learning (RL) task that balances immediate user responses with longer-term outcome objectives. Building on the original Decision Transformer architecture ~\cite{chen2021decisiontransformerreinforcementlearning}, we propose a trajectory formulation that enables the model to infer return-to-go (RTG) distributions from historical states, using the predicted RTG as a prompt for action selection. We also introduce a prompt tuning strategy, which provides direct control over the reward-seeking behavior of the policy and supports the generation of diverse notification strategies suited to different operational goals.

To support large-scale, near-real-time serving in a production recommender system (150K QPS), we design a circular buffer mechanism for efficient sequence feature persistence and updates, along with a tenant-specific environment for cloud-based model hosting. Extensive offline evaluations and online A/B testing indicate improvements in notification relevance and overall session activity, achieving a +0.72\% increase in user sessions compared to a CQL-based baseline.

For future work, we plan to investigate architectural extensions, including hybrid Decision Transformer and Q-learning methods ~\cite{yamagata2023qlearningdecisiontransformerleveraging}, critic-guided Decision Transformers ~\cite{wang2023criticguideddecisiontransformeroffline}, longer temporal context windows, and refined reward modeling strategies.
\begin{acks}
Thanks to all the partners (Xiaoxiao Guo, Yichen Tu, Tianqi Wang, Lichen Wang, Chunan Zeng, Daniel Lau, Zhuoran Yu, Nirav Shingala, Charles Xiao, Keerthi Selvaraj) who collaborated with us to build and scale an effective notification decision making system at LinkedIn. We also thank Angus Qiu, Aman Gupta, Ajay Vaddadi, Mohsen Jamali, Swetha Nagabhushan Karthik, Dawn Woodard, Xiaobing Xue and Julie Choi for leadership support.
\end{acks}

\bibliographystyle{ACM-Reference-Format}
\bibliography{sample-base}





\end{document}